\documentclass[twocolumn]{article}
\usepackage{geometry}           
\usepackage[compact]{titlesec}  
\usepackage{graphicx}           
\usepackage{subcaption}         
\usepackage{float}              
\usepackage{booktabs, multirow, tabularx, array, colortbl}
\usepackage{adjustbox}          
\usepackage{pdflscape}          
\usepackage{rotating}           
\usepackage{amsmath, amssymb}   
\usepackage{xcolor}             
\usepackage{sectsty}            
\usepackage{hyperref}           
\usepackage{authblk}            
\usepackage{lipsum}             
\usepackage{cite}
\usepackage{parskip}
\usepackage{enumitem}

\title{\textbf{An Intuitionistic Fuzzy Logic Driven UNet architecture: Application to Brain Image segmentation}}

\author[1]{Hanuman Verma\thanks{Corresponding author: \href{mailto:hv4231@gmail.com}{hv4231@gmail.com}}}
\author[2,3]{Kiho Im}
\author[4,6]{Pranabesh~Maji}
\author[5,6]{Akshansh Gupta}

\affil[1]{Department of Mathematics, Bareilly College, Bareilly (MJP Rohilkhand University), Uttar Pradesh, India}
\affil[2]{Division of Newborn Medicine, Fetal Neonatal Neuroimaging and Developmental Science Center, Boston Children’s Hospital, Harvard Medical School, Boston, MA 02115, USA}
\affil[3]{ Department of Pediatrics, Harvard Medical School, Boston, MA, USA }
\affil[4]{CSIR–Central Electronics Engineering Research Institute, Pilani 333031, Rajasthan, India}
\affil[5]{CSIR–National Institute of Science Communication and Policy Research, New Delhi, India}
\affil[6]{Academy of Scientific and Innovative Research (AcSIR), Ghaziabad 201002, India}

\date{}  
\begin{document}
\maketitle
	
\begin{abstract}
Accurate segmentation of MRI brain images is essential for image analysis, diagnosis of neuro-logical disorders and medical image computing. In the deep learning approach, the convolutional neural networks (CNNs), especially UNet, are widely applied in medical image segmentation. However, it is difficult to deal with uncertainty due to the partial volume effect in brain images. To overcome this limitation, we propose an enhanced framework, named UNet with intuitionistic fuzzy logic (IF-UNet), which incorporates intuitionistic fuzzy logic into UNet. The model processes input data in terms of membership, non-membership, and hesitation degrees, allowing it to better address tissue ambiguity resulting from partial volume effects and boundary uncertainties. The proposed architecture is evaluated on the Internet Brain Segmentation Repository (IBSR) dataset, and its performance is computed using accuracy, Dice coefficient, and intersection over union (IoU). Experimental results confirm that IF-UNet improves segmentation quality with handling uncertainty in brain images.
\end{abstract}
	
	\vspace{1em}

\section{Introduction}\label{intro}
Automatic segmentation of brain tissues from magnetic resonance imaging (MRI) scans, specifically cerebrospinal fluid (CSF), gray matter (GM), and white matter (WM), plays a crucial role in medical image computing. The accurate segmentation of brain tissues such as WM, GM and CSF has significant clinical relevance, as these tissues serve as essential biomarkers for diagnosing and monitoring neurological disorders. It allows clinicians to detect subtle structural changes associated with diseases such as multiple sclerosis, Alzheimer’s disease, and brain atrophy. It also enhances the reliability of volumetric measurements used to track disease progression and evaluate treatment response. Furthermore, precise tissue delineation reduces clinician variability, supports improved surgical and therapeutic planning, and contributes to more consistent, patient-specific clinical decision-making in neuroimaging practice.	A wide range of segmentation methods,  supervised and unsupervised, hybrid metaheuristic–clustering approaches, active-contour models, subject-specific atlas techniques, and graph-cut methods have been developed to medical image segmentation \cite{Verma2016IFCM,Gupta2023Book,Fawzi2021Review,Wang2022Survey,jafrasteh2024enhanced,Verma2025KIFECM,soloh2024brain}. In recent years, deep learning approaches, particularly convolutional neural networks (CNNs), a prominent part of machine learning, have gained significant attention for brain image segmentation \cite{moeskops2016automatic}. Among these, the UNet architecture \cite{Ronneberger2015UNet} and its several extensions, including Attention UNet \cite{Oktay2018AttentionUNet}, Residual UNet \cite{Zhang2018DRUNet}, and 3D UNet \cite{Cicek2016UNet3D}, have been introduced to improve feature extraction, localisation, and segmentation accuracy. The original UNet framework, proposed by Ronneberger et al. \cite{Ronneberger2015UNet}, consists of a contracting path to capture contextual information and an expansive path for precise localisation. Over time, multiple improved variants of UNet have been developed to enhance the segmentation performance  further \cite{Azad2024UNetReview,Punn2022Survey,Liu2020SurveyUnet}.

In human brain MR images, uncertainty in images often arises due to intensity variations and transitional regions between different tissues, making the accurate segmentation of pixels into their actual tissue classes a challenging task. Moreover, various types of artifacts may occur during MRI scanning \cite{Henkelman1985MR}. A notable artifact is the partial volume effect, in which a single voxel may contain contributions from multiple tissue types, resulting in ambiguous intensity values. This issue is prevalent at tissue boundaries and significantly declines segmentation accuracy. To address such challenges, intuitionistic fuzzy logic, derived from intuitionistic fuzzy set (IFS) theory \cite{Atanassov1986IFS} that extends classical fuzzy set theory \cite{Zadeh1965Fuzzy}, introduces an additional component of uncertainty, termed hesitation, that enhances decision-making in the presence of ambiguous information. Among the unsupervised clustering methods using IFS theory, the intuitionistic fuzzy c-means (IFCM) algorithm and its various variants have been widely applied in medical image segmentation \cite{kumar2019modified,Verma2016IFCM,Verma2025KIFECM}. 

Recently, non–deep-learning methods for brain image segmentation include fuzzy-clustering variants based on fuzzy and intuitionistic fuzzy logic, hybrid metaheuristic–clustering approaches, active-contour models, subject-specific atlas techniques, and graph-cut methods. Fuzzy c-means and its kernel or intuitionistic extensions improve noise robustness and soft-boundary handling. However they still rely heavily on hand-tuned spatial parameters and face intensity inhomogeneity and heterogeneous lesions \cite{jafrasteh2024enhanced,Verma2025KIFECM,Verma2016IFCM}. Atlas-based and subject-specific registration methods can generate anatomically accurate tissue labels, however, their performance degrades with large anatomical variability or registration errors \cite{noorizadeh2024subject}. Level-set and graph-cut approaches effectively encode geometric priors and produce accurate contours in high-contrast images, but they require manual initialisation, per-case parameter tuning, and problem-specific energy functions that do not generalise well across modalities \cite{soloh2024brain}

Although, the majority of brain image segmentation methods utilise CNNs, specifically the UNet. Sometimes the UNet is incapable of addressing uncertainty arising from ambiguity and the effect of the partial volume effect. In the literature, some studies have made efforts to use fuzzy logic in deep learning to mitigate this uncertainty. Price et al.\cite{Price2019FuzzyLayers} introduced the inclusion of a fuzzy layer in deep learning utilising the Choquet and Sugeno fuzzy integrals. To extract valuable information, Sharma et al. \cite{Sharma2019Pooling} added a unique fuzzy logic-based pooling layer to the convolutional layer, where the distinctive pooling layer utilizes fuzzy logic to extract information using two separate approaches and evaluated for image classification. Ding et al. \cite{Ding2021Infant} proposed a multimodal framework employing deep learning integrated with fuzzy logic, termed as fuzzy-informed deep learning segmentation (FI-DL-Seg), for the segmentation of newborn brain MRI. This framework supports input images as T1 and T2-weighted images. To address the ambiguity in convolutional feature values, the FI-DL-Seg model incorporates fuzzy and uncertainty layers, volumetric fuzzy pooling, and fuzzy-enabled multiscale learning. These characteristics aid in the information extraction process from an ambiguous MRI brain image. Fuzzy logic in convolutional neural networks was proposed by Huang et al.\cite{Huang2021BreastSeg} for use in ultrasound breast image segmentation. The image is initially enhanced by wavelet transform, followed by its conversion into the fuzzy domain using diverse membership functions to address the uncertainty present in image. For the semantic segmentation of breast cancers from ultrasound images, Badawy et al. \cite{Badawy2021Breast} suggested a deep learning model that integrates fuzzy logic with CNNs. Here, fuzzy logic is used for pre-processing, and CNNs are used for segmentation. Huang et al.\cite{Huang2022Trustworthy} suggested a technique that utilises fuzzification to minimise the amount of uncertainty in CNNs. This technique is intended to segregate breast images from ultrasound images. It aims to reduce uncertainty in image segmentation and is referred to as the spatial and channel-wise fuzzy uncertainty reduction network. A three-way decision-making process is employed by Subhashini et al.\cite{Subhashini2022ThreeWay} to address the uncertainty problem that arises from vagueness by combining fuzzy logic with deep learning models. The target-aware UNet model with fuzzy skip connections was introduced and used for pancreatic segmentation by Chen et al.\cite{Chen2022Pancreas} that integrates fuzzy logic with the UNet architecture. The decoder component of this model also includes a target mechanism to improve feature representation for the targeted segmentation. Fuzzy attention neural networks, which combine attention and fuzzy logic in a 3D UNet architecture, were introduced by Nan et al. \cite{Nan2023FuzzyAttention}. It detected airway discontinuities and decreased the uncertainty associated with airway segmentation from computed tomography images. In order to address boundary uncertainty in medical image segmentation, Zhou et al.\cite{zhou2025f2cau} introduced the F2CAU-Net architecture, a dual fuzzy medical image segmentation cascade method that integrates fuzzy logic and convolutional operations to extract both local and global features within a deep learning framework. This approach helps improve segmentation performance in regions with ambiguous or unclear boundaries. A review of fuzzy deep learning for uncertain medical data has been presented in the literature \cite{zheng2024systematic}. It discusses the integration of fuzzy logic into deep learning to address various types of uncertainty in medical data, as compared to crisp values in the classical set theory. Zhang and Xue \cite{zhang2025fuzzy} introduced fuzzy attention-based deep neural networks that incorporate fuzzy logic and attention mechanisms. This work includes a fuzzy squeeze-and-excitation densely connected convolutional network with an orthogonal projection loss for the precise diagnosis of Acute Lymphoblastic Leukemia. The experimental results show improved accuracy in the diagnosis.

The UNet architecture and its variants rely on crisp decision boundaries derived from classical set theory. Such an approach is not able to capture the uncertainty or hesitation that occurs at the boundary of images, where pixel intensities are mixed due to partial volume effects. Addressing this limitation requires an uncertainty-aware or fuzzy approach that can effectively model the transitions between regions.

In this work, to integrate the intuitionistic fuzzy logic into the baseline UNet architecture, we have presented UNet with intuitionistic fuzzy logic (IF-UNet) to handle the uncertainty caused by the partial volume effect in brain images. The main contribution in the proposed framework is:

\begin{enumerate} [label=\roman*)]
	\item In the contraction (down-sampling) path of IF-UNet, the input image is processed in its intuitionistic fuzzy representation, including additional layers corresponding to membership, non-membership, and hesitation degrees.
	\item The hesitation degree is incorporated to handle uncertainties arising from vague or ambiguous tissue boundaries.
	\item The fuzzy-based input representations help the network extract additional uncertainty-related features, thereby improving segmentation performance.
	
\end{enumerate}

The proposed IF-UNet, baseline UNet, and Attention UNet architectures are trained on publicly available Internet Brain Segmentation Repository (IBSR) dataset, and evaluated their segmentation performance in terms of accuracy (AC), Dice coefficient (DC), and intersection over union (IoU). Experimental results confirm that IF-UNet improves segmentation quality by handling uncertainty in brain images.

The rest of this paper is organised as follows: Section 2 provides an overview of intuitionistic fuzzy sets and their intuitionistic representation of data. Section 3 details the proposed IF-UNet architecture. Section 4 outlines the experimental setup, and Section 5 reports the results along with their analysis. Finally, Section 6 offers the concluding remarks.

\section{Intuitionistic fuzzy set and intuitionistic representation of data}\label{ifs} 

The concept of an \textit{intuitionistic fuzzy set} (IFS), introduced by Atanassov \cite{Atanassov1986IFS}, is defined on a universe set $X$ using a membership function $\mu_A:X\to[0,1]$ and a non-membership function $\nu_A:X\to[0,1]$. Mathematically, it is written as:  
\begin{equation}
A = \left\{ \langle x, \mu_A(x), \nu_A(x) \rangle : \forall x \in X \right\}
\end{equation}

In IFS, membership and non-membership degrees satisfy the constraint  
$0 \leq \mu_A(x) + \nu_A(x) \leq 1$. The hesitation degree, denoted by $\pi_A(x)$, measures the uncertainty and is computed as $\pi_A(x) = 1 - \mu_A(x) - \nu_A(x).$  When the hesitation value becomes zero, the IFS reduces to a fuzzy set \cite{Zadeh1965Fuzzy} and each element either belongs or does not belong to the set. An IFS is thus characterised by three components: membership, non-membership, and hesitation degrees. The membership degree is often computed using functions such as triangular, trapezoidal, Gaussian, or sigmoid types. The non-membership degree can be computed through negation functions discussed in the literature \cite{Bustince2000Generators}, such as those by Sugeno \cite{Bustince2000Generators} and Yager \cite{Yager1980Fuzziness}. Using Sugeno’s negation function, the non-membership degree is computed \cite{Bustince2000Generators} as $N(\mu(x)) = \frac{1 - \mu_A(x)}{1 + \lambda \mu_A(x)}, \quad \lambda > 0$. The IFS can be rewritten as: 
\begin{equation}
A_{\text{Sugeno}} = \left\{ \langle x, \mu_A(x), \nu_A(x) = \tfrac{1 - \mu_A(x)}{1 + \lambda \mu_A(x)} \rangle : \forall x \in X \right\}
\end{equation}

Further, intuitionistic fuzzification of an image refers to converting crisp pixel values into an intuitionistic fuzzy representation that allows interpretations for uncertainty and vagueness. For an image $B = \{x_j\}_{j=1}^N \quad \text{with $N$ pixels},$  
the transformed form is written as:  
\begin{equation}
B^{IFS} = \{x_j^{IFS}\}_{j=1}^N
\end{equation}
where each pixel is represented as $x_j^{IFS} = \big(\mu_B(x_j), \nu_B(x_j), \pi_B(x_j)\big)$.  
Here, $\mu_B(x_j)$, $\nu_B(x_j)$ and $\pi_B(x_j)$ denote the membership, non-membership, and hesitation degrees, respectively.

\begin{figure}[h]
	\centering
	\begin{subfigure}{0.12\textwidth}
		\includegraphics[scale=0.23]{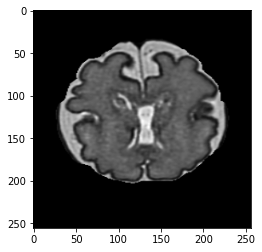}
		\caption{}
	\end{subfigure}
	\begin{subfigure}{0.12\textwidth}
		\includegraphics[scale=0.23]{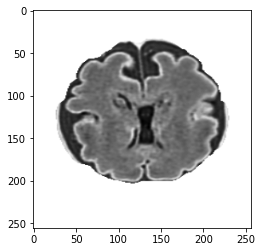}
		\caption{}
	\end{subfigure}
	\begin{subfigure}{0.12\textwidth}
		\includegraphics[scale=0.23]{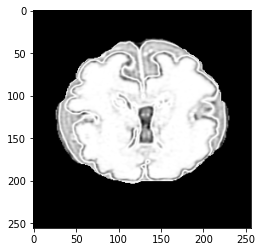}
		\caption{}
	\end{subfigure}
	
	\caption{Intuitionistic fuzzy data in form of (a): membership, (b): non-membership and (c): hesitation degree. }
	\label{fig:Fig1}
\end{figure}

An image is converted into an intuitionistic fuzzy form, denoted as $B^{IFS}$ to handle uncertainty in transition regions, thereby enabling better pixel classification. Image features are represented by membership, non-membership, and hesitation degrees. The CNNs for image classification utilise these intuitionistic fuzzy representations for improved feature extraction. The hesitation component highlights boundary regions, offering additional information for feature analysis. Higher hesitation values indicate more uncertainty, while zero hesitation reflects clear membership or non-membership. The membership and non-membership values define the degree of belonging or non-belonging for each pixel, and their distributions are illustrated in Fig.\ref{fig:Fig1}.

\section{UNet with Intuitionistic Fuzzy Logic (IF-UNet)}\label{Sec3}

The UNet model, introduced by Ronneberger et al. \cite{Ronneberger2015UNet}, is a variant of a convolutional neural network developed for biomedical image segmentation. Its structure resembles the English letter “U,” consisting of two main components: a contracting path (down-sampling) that extracts significant features and an expanding path (up-sampling) that enables accurate localization. In the contracting path, convolution layers and max-pooling operations are used. In contrast the expanding path reconstructs spatial details pixel by pixel and integrates them with features from the contracting stage to retain essential information. The UNet has proven highly effective for medical image segmentation, delivering reliable and precise results. However, the UNet framework, despite leveraging deep learning, has challenges in handling uncertainties at boundaries.


\begin{figure*}[!t] 
	\centering
	\includegraphics[scale=0.60]{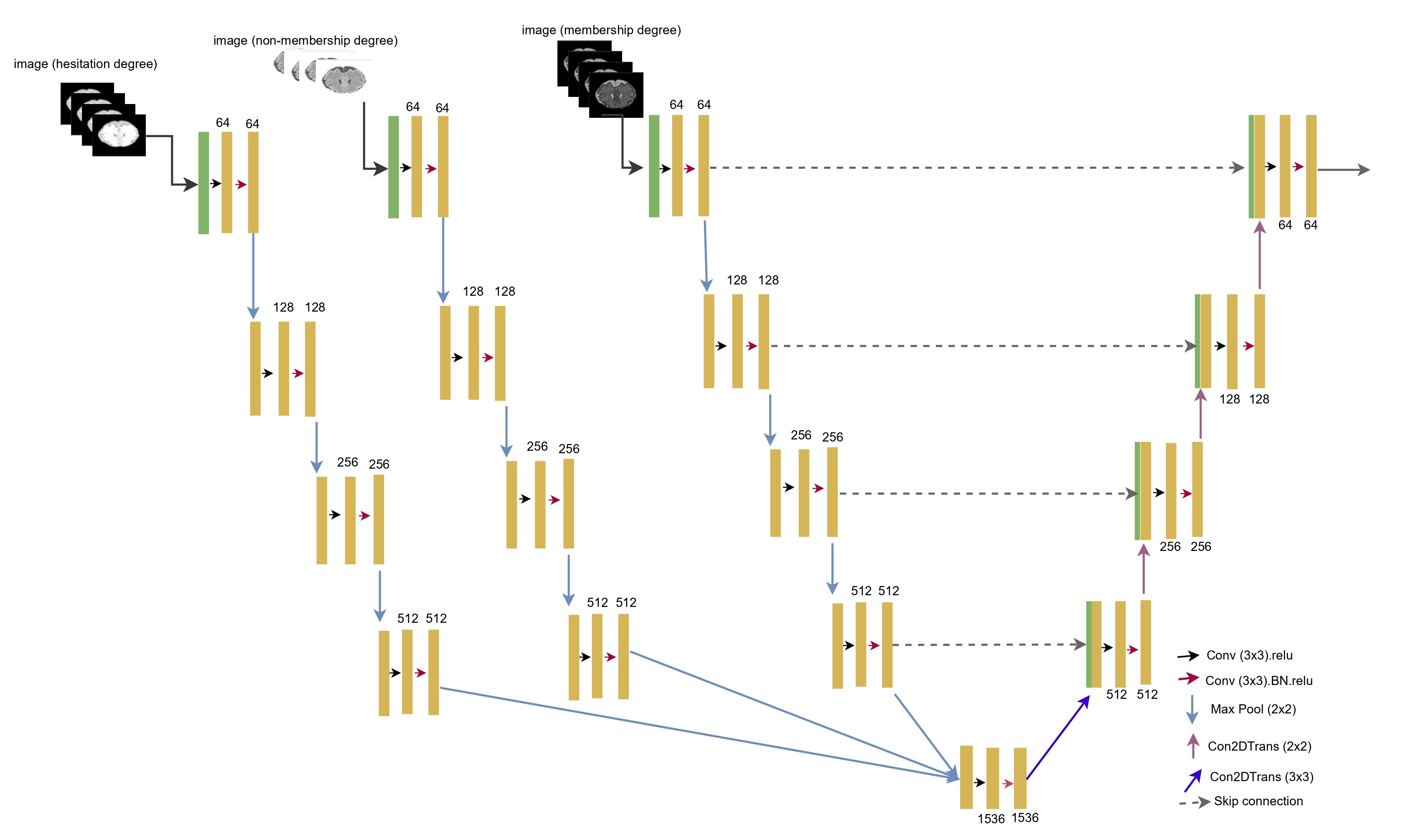}
	\caption{A framework for IF-UNet architecture. It processes the input image in terms of intuitionistic fuzzification, where the image is represented in term of membership, non-membership, and hesitation degree. The encoder part processes the data in terms of membership and non-membership, while the decoder part follows the UNet architecture.}
	\label{fig:Fig2}
\end{figure*}
To integrate intuitionistic intuition and extract vagueness information, we incorporate intuitionistic fuzzy sets, and design the architecture, named UNet with intuitionistic fuzzy logic (IF-UNet). In IF-UNet, the feature of input data is accepted in the intuitionistic fuzzy form, with converting the image into intuitionistic fuzzy representation form that is 
 $x_j^{\mathrm{IFS}} = \big(\mu_B(x_j), \nu_B(x_j), \pi_B(x_j)\big).$ 
 These terms represent the membership degree $\mu_B(x_j)$,non-membership degree $\nu_B(x_j)$, and hesitation degree $\pi_B(x_j)$, as shown in Fig.~\ref{fig:Fig2}. Each pixel of the image is converted into qualitative features through intuitionistic fuzzy encoding. The intuitionistic fuzzy representation of data extracts features across different intensity levels. The hesitation part captures the uncertainty features across various boundaries in the brain image. The uncertainty in IF-UNet can be handled using intuitionistic fuzzy logic. This involves transforming the data into a three-way representation, represented as a vector consisting of membership, non-membership, and hesitation degrees. Stacking these input data in the form of intuitionistic fuzzification converts the image into a set of relevant feature vectors. Additionally, the skip connection in the up-sampling part preserves the information. This approach helps to handle boundary pixels, which shows the vagueness, as the data is interpreted through an intuitionistic fuzzy representation form. Membership value indicates that pixels belong to a similar group, non-membership degree indicates that pixels belong to a different group, and hesitation degree determines the assignment of boundary pixels to a particular group, aiding in deciding the classification of pixels. The up-sampling part examines the location of the pixels, and intuitionistic logic assists in decision-making. Intuitionistic fuzzy logic improves decision-making and classifies the pixels into the desired classes.

\section{Experiment}\label{Sec4}
\subsection{Dataset }\label{Sec4.1}
In this work, we take the IBSR (Internet Brain Segmentation Repository) data for the validation of the proposed framework. The IBSR dataset, available at \url{https://www.nitrc.org/projects/ibsr/} is a publicly available dataset to evaluate brain image segmentation methods, as it provides manually segmented MRI data. It consists of 20 T1-weighted MRI scans of healthy subjects and each is accompanied by expert-generated segmentation labels. These detailed annotations over various brain tissues and structures make for reliable ground truth data. Well-structured labelling is essential for training segmentation models and objectively validating the performance of the architecture.
\subsection{Baseline and implementation }\label{Sec4.2}
To implement the proposed IF-UNet approach, we transformed the brain data into an intuitionistic fuzzy representation $x_j^{\mathrm{IFS}} = \big(\mu_B(x_j), \nu_B(x_j), \pi_B(x_j)\big)$ utilising Sugano’s negation functions. The model accepts input in the form of data, corresponding to membership, non-membership and hesitation degree, allowing it to better capture uncertainty-related information.

To evaluate segmentation performance, the proposed IF-UNet, UNet and Attention UNet architectures are trained on 10 cases of IBSR MRI brain data (2,560 images) and compared their segmentation performance along with validation. The categorical cross-entropy as the loss function, with ReLU activation and the Adam optimizer function are used during the training of the model. Batch normalization and dropout layers are incorporated after each convolutional block to reduce overfitting, while the convolution operations use a kernel size of $3 \times 3$. The network configuration detail is presented in Fig.~\ref{fig:Fig2}. The architectures are implemented in Keras with a TensorFlow backend. Performance is quantitatively assessed using standard metrics such as accuracy $(AC)$, Dice coefficient $(DC)$, and intersection over union $(IoU)$, along with the validation results such as accuracy validation $(AC\_val)$, Dice coefficient validation $(DC\_val)$ and intersection over union validation $(IoU\_val)$. In training, we set the number of epochs to 100 and the batch size to 2 for the IBSR data. For dataset partitioning, an 80:20 split is used, with $80\%$ of data used for training and the remaining $20\%$ reserved for testing.

\section{Results and discussion}\label{Sec5}

\begin{figure}[h!]
	\begin{subfigure}[b]{0.1\textwidth}
		\includegraphics[scale=0.4]{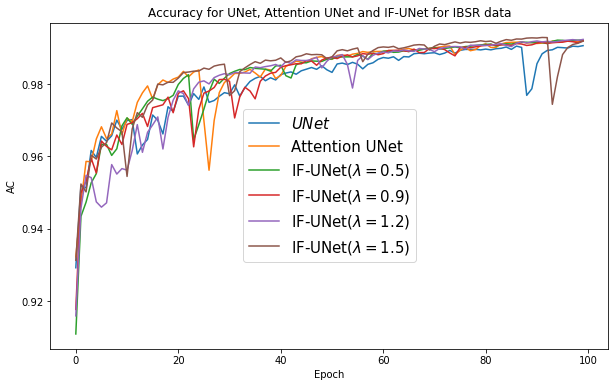}
		\caption{}
		\label{fig:3a}
	\end{subfigure}
	
	\vspace{0.5cm}  
	
	\begin{subfigure}[b]{0.1\textwidth}
		\includegraphics[scale=0.4]{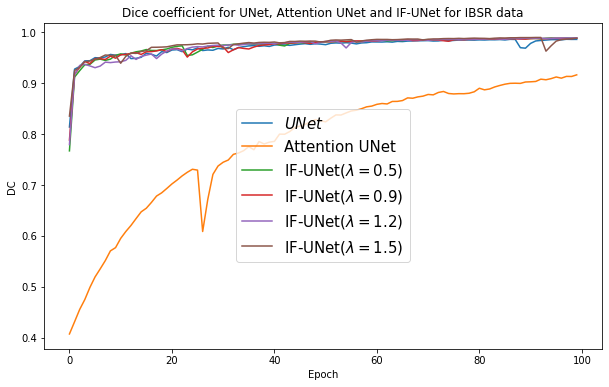}
		\caption{}
		\label{fig:3b}
	\end{subfigure}
	
	\vspace{0.5cm}  
	
	\begin{subfigure}[b]{0.1\textwidth}
		\includegraphics[scale=0.4]{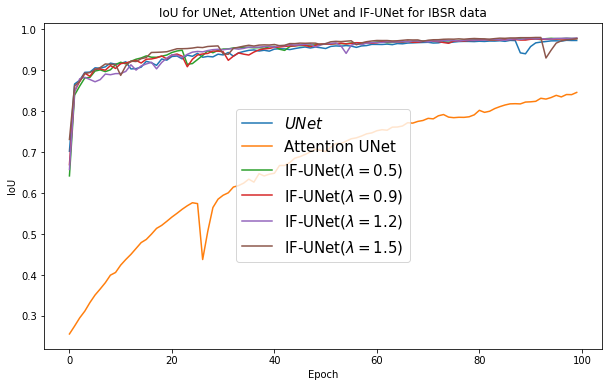}
		\caption{}
		\label{fig:3c}
	\end{subfigure}
	
	\caption{The training performance of UNet, Attention UNet and IF-UNet during the training process in terms of $AC$,$DC$,and $IoU$ are shown in Fig.~\ref{fig:Fig3}\subref{fig:3a}--\subref{fig:3c} respectively. The various value of $\lambda(=0.5,0.9,1.2,1.5)$ in the Sugeno negation function is considered in the study for IF-UNet architectures, and it show better results in comparison to UNet and Attention UNet. It can be shown that the segmentation performance of IF-UNet varies with different $\lambda$ values, yet consistently better performance in comparison to UNet and Attention UNet.}
	\label{fig:Fig3}
\end{figure}

An ablation study of the proposed IF-UNet architecture for MRI brain image segmentation was carried out by integrating intuitionistic fuzzy logic with Sugeno negation functions into the CNNs framework. The baseline UNet and Attention UNet architectures with categorical cross-entropy loss, are used for comparison. The evaluation was performed on the IBSR dataset, which consists of $20$ T1-weighted MRI brain scans along with their corresponding expert-labelled segmentation masks. The segmentation performance of IF-UNet, baseline UNet and Attention UNet architectures is shown in Fig.~\ref{fig:Fig3} in terms of $AC$,$DC$ and $IoU$ during the training over $100$ epochs. The IF-UNet tested for various values of $\lambda$, and the results show that the IF-UNet architecture also depends on the value of $\lambda$ in the Sugeno negation function. 

\begin{table}[ht]
	\centering
	\caption{A quantitative evaluation of segmentation performance for UNet, Attention UNet and IF-UNet, which incorporates intuitionistic fuzzy logic with the Sugeno negation function, across different values of $\lambda$ for the IBSR dataset. The training was performed for 100 epochs with a batch size of 2. The effectiveness of segmentation was assessed using three standard metrics: accuracy $(AC)$, Dice coefficient $(DC)$, and intersection over union $(IoU)$ and corresponding validations that accuracy validation $(AC\_val)$, Dice coefficient validation $(DC\_val)$, and intersection over union validation $(IoU\_val)$}.
	\label{tab:Table1}
	\resizebox{0.98\linewidth}{!}{
		\begin{tabular}{lcccccc}
			\hline
			\textbf{Metrics} & \textbf{UNet} & \textbf{Attention} & \multicolumn{4}{c}{\textbf{IF-UNet}} \\
			\cline{4-7}
			& & \textbf{UNet}& $\lambda=0.5$ & $\lambda=0.9$ & $\lambda=1.2$ & $\lambda=1.5$ \\
			\hline
			AC       & 0.9907 & 0.9919 & 0.9922 & 0.9921 & 0.9924 & 0.9919 \\
			DC       & 0.9861 & 0.9165 & 0.9886 & 0.9886 & 0.9892 & 0.9884 \\
			IOU      & 0.9728 & 0.8459 & 0.9776 & 0.9776 & 0.9788 & 0.9772 \\
			AC\_val  & 0.9886 & 0.9898 & 0.9904 & 0.9910 & 0.9912 & 0.9900 \\
			DC\_val  & 0.9841 & 0.9211 & 0.9867 & 0.9848 & 0.9879 & 0.9850 \\
			IOU\_val & 0.9688 & 0.8543 & 0.9739 & 0.9703 & 0.9762 & 0.9705 \\
			\hline
		\end{tabular}
	}
\end{table}

\begin{figure}[H]  
	\includegraphics[scale=0.45]{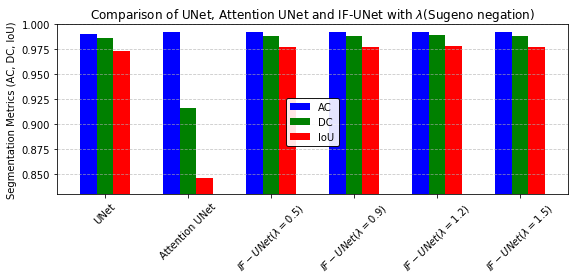}
	
	\caption{The comparison of segmentation performance of UNet, Attention UNet and the proposed IF-UNet with Sugeno negation function, measured in terms of $AC, DC$, and $IoU$ on the IBSR dataset using different values of $\lambda$ in the Sugeno negation function. The experimental results show that the proposed IF-UNet consistently achieves improved segmentation accuracy across all tested values of $\lambda$, demonstrating its effectiveness in capturing vague and qualitative information compared to UNet and Attention UNet.}
	\label{fig:Fig4}
\end{figure}

\begin{figure}[H]  
	\includegraphics[scale=0.45]{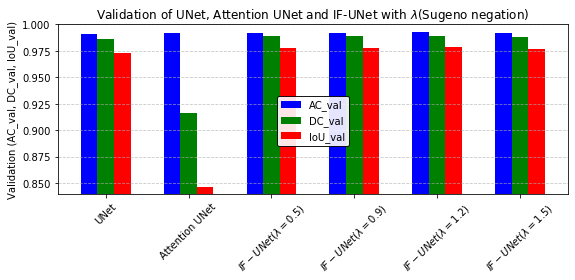}
	
	\caption{The comparison of training validation of UNet, Attention UNet and IF-UNet with the Sugeno negation function, measured in terms of $AC\_val,DC\_val$, and $IoU\_val$ on IBSR dataset for various of $\lambda (=0.5,0.9,1.2,1.5)$ in Sugeno negation function. In training validation, the results show that the IF-UNet consistently achieves improved segmentation validation in comparison to UNet and Attention UNet.}
	\label{fig:Fig5}
\end{figure}

The quantitative segmentation results with 100 epochs are summarized in Table~\ref{tab:Table1} in terms of $AC$,$DC$, $IoU$ and respective validation that is $AC\_{val}$,$DC\_{val}$,$IoU\_{val}$. By incorporating intuitionistic fuzzy logic with the Sugeno negation function, IF-UNet effectively addresses uncertainty and enhances segmentation accuracy for IBSR images. For various values of $\lambda$, the method achieved consistent and reliable results on the IBSR dataset. It can be observed that at $\lambda=1.2$, the proposed framework IF-UNet achieves the highest segmentation performance in terms of $AC = 0.9924, DC = 0.9892$, and $IoU = 0.9788$ for the IBSR dataset. The comparative performance is further illustrated through the bar charts in Fig.~\ref{fig:Fig4} and Fig.~\ref{fig:Fig5}, which clearly indicate that the proposed approach IF-UNet performs better  in comparison to baseline UNet and Attention UNet. The Attention U-Net shows relatively poor performance on the IBSR dataset.

\begin{figure*}[!t]  
\centering
\includegraphics[scale=0.1]{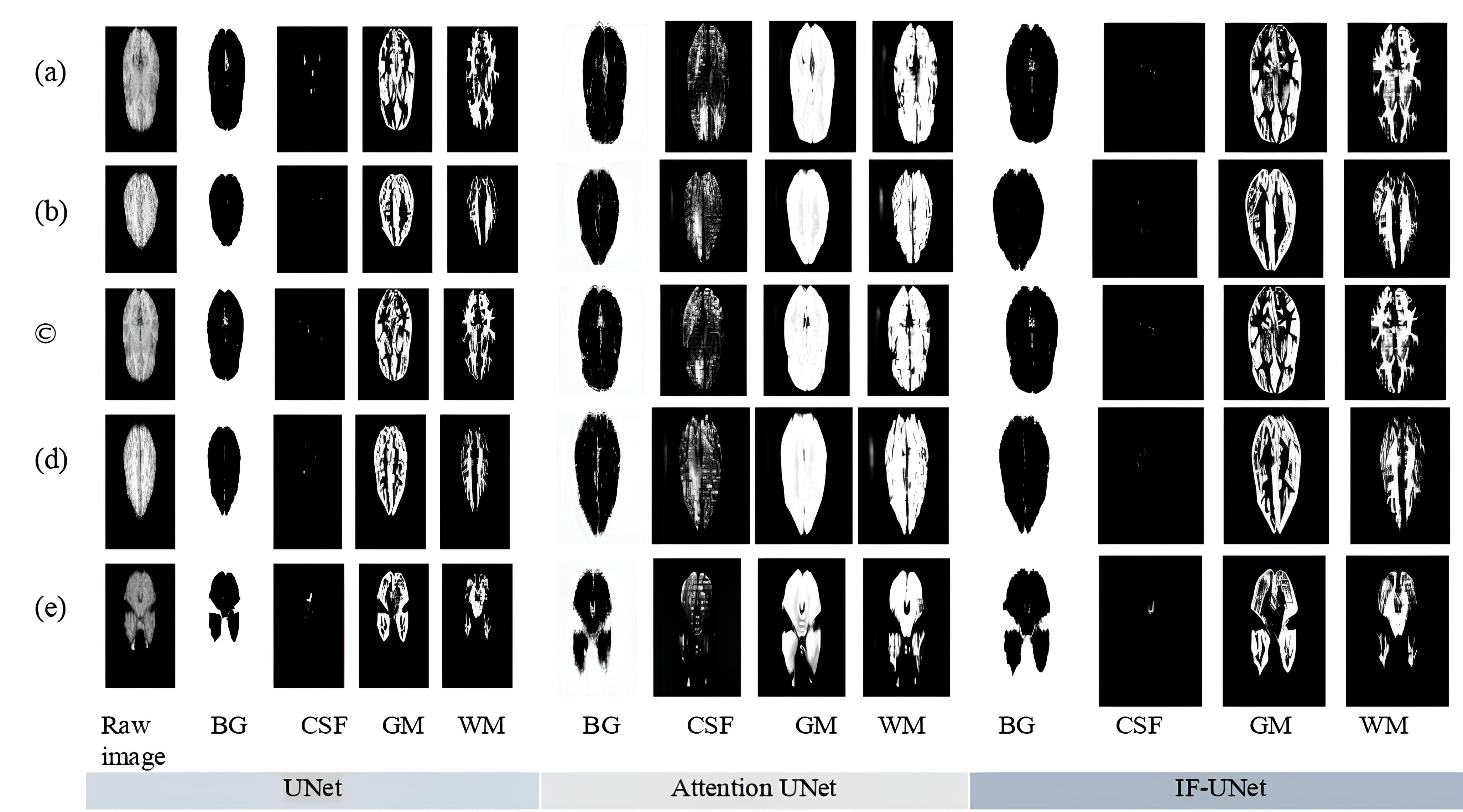}
\caption{Illustration of brain MRI slice segmentation into different classes, including BG, CSF, GM, and WM, using UNet, Attention UNet and IF-UNet. The figure presents results from UNet (left side), Attention UNet (middle) and IF-UNet (right side) at $\lambda (=0.5)$. Incorporating intuitionistic fuzzy logic into these architectures effectively addresses the partial volume effect, leading to improved segmentation outcomes.}
\label{fig:Fig6}
\end{figure*}

The trained IF-UNet, Attention UNet, and UNet architectures on the training data are used for segmenting the raw 2D data into various classes: background (BG), CSF, GM, and WM. The CSF, GM, and WM are the brain tissues, and the BG is the background of the 2D brain image. The 5 different 2D raw images are segmentated into the four classes: BG, CSF, GM, and WM as represented in Fig.~\ref{fig:Fig6}. The Fig.~\ref{fig:Fig6}, show the visual comparison result of IF-UNet, UNet and Attention UNet architectures. The quantitative results indicate that the segmentation performance of IF-UNet achieves better outcomes compared to baseline UNet and Attention UNet architectures. An ablation study was also conducted to analyze the contribution of intuitionistic fuzzy logic, showing that incorporating fuzzy-based information significantly enhances segmentation quality. Unlike the baseline UNet and Attention UNet, the proposed IF-UNet framework effectively exploits intuitionistic fuzzy set theory to improve segmentation accuracy. Although the incorporation of intuitionistic fuzzy logic improves robustness to uncertainty, structural abnormalities may still affect the accuracy of boundary detection and tissue classification. In the IF-UNet architecture, segmentation errors may occur in highly deformed or excessively noisy brain images.
\begin{table}[h!]
	\centering
	\caption{Computational complexity of architectures}
		\label{tab:Table2}
	\begin{tabular}{lcc}
		\hline
		Architectures & Trainable & Inference time \\
		& parameters&(sec/image)\\ 
		\hline
		UNet & 31,036,676 & 0.7244 \\
		Attention UNet & 37,330,576 & 0.6165 \\
		IF-UNet ($\lambda = 0.5$) & 73,703,300 & 88.5058 \\
		\hline
	\end{tabular}
\end{table}
\\
The computational complexity of UNet, Attention UNet, and the proposed IF-UNet is evaluated using their trainable parameters and average inference time per image, shown in Table~\ref{tab:Table2}. Each architecture was run 50 times on the same image and computed the average inference time. From Table~\ref{tab:Table2}, it is observed that IF-UNet, which integrates intuitionistic fuzzy representations along with additional processing layers, has a significantly higher trainable parameters and average time, reflecting the extra computational effort required to process membership, non-membership, and hesitation degree components. In medical imaging, accuracy is more important than speed, as clinical decisions depend on precise and reliable segmentation. Although IF-UNet requires more computation time, its improved segmentation quality over UNet and Attention UNet makes it more suitable for applications where dependable results are essential for patient care.

\section{Conclusion }\label{Sec6}
To address uncertainty in feature representation in biomedical image segmentation, we propose a framework, named IF-UNet, which integrates intuitionistic fuzzy logic into the UNet and Attention UNet architecture. In IF-UNet, input data is processed in intuitionistic fuzzy form, allowing more effective distinction between various tissues: WM, CSF, GM and boundary regions. By leveraging uncertainty information during training, IF-UNet improves segmentation accuracy. The IF-UNet, baseline UNet and Attention UNet architectures are trained on IBSR brain data and evaluated segmentation performance in terms of AC,DC, and IoU and corresponding validation $AC\_val$,$DC\_val$, and $IoU\_val$. Both ablation studies and experimental results demonstrate that IF-UNet consistently performs better than the baseline UNet and Attention UNet, particularly across different parameter values in the Sugeno negation function.
\vspace{-5mm}
\section*{Author contributions}
Hanuman Verma: Conceptualization, Methodology, Validation, Data preparation, Writing– original draft. Kiho Im: Investigation, review \& editing, Supervision. Pranabesh Maji: Validation, Methodology, reviewing. Akshansh Gupta: Methodology, review \& editing. All authors reviewed the manuscript.
\vspace{-5mm}
\section*{Data Availability and Materials}
This study used the publicly available Internet Brain Segmentation Repository (IBSR) dataset that can be accessed from \url{https://www.nitrc.org/projects/ibsr/} 
\vspace{-5mm}
\section*{Declarations}
Conflict of interest: The authors declare no competing interests.
\vspace{-5mm}
\section*{Funding}
This research project is supported by the Innovative Research Grant (IRG), file No: IRG/MJPRU/DoR/2022/06, under the Directorate of Research, MJP Rohilkhand University, Bareilly, Uttar Pradesh, India. 

\bibliographystyle{plain}
\bibliography{sn-ref-shorting}

@article{Verma2016IFCM,
  author    = {Verma, H. and Agrawal, R. K. and Sharan, A.},
  title     = {An improved intuitionistic fuzzy c-means clustering algorithm incorporating local information for brain image segmentation},
  journal   = {Applied Soft Computing},
  volume    = {46},
  pages     = {543--557},
  year      = {2016}
}

@book{Gupta2023Book,
  editor    = {Gupta, A. and Verma, H. and Prasad, M. and Kirar, J. S. and Lin, C. T.},
  title     = {Computational Intelligence Aided Systems for Healthcare Domain},
  publisher = {CRC Press},
  year      = {2023}
}

@article{Fawzi2021Review,
  author    = {Fawzi, A. and Achuthan, A. and Belaton, B.},
  title     = {Brain image segmentation in recent years: A narrative review},
  journal   = {Brain Sciences},
  volume    = {11},
  number    = {8},
  pages     = {1055},
  year      = {2021}
}

@article{Wang2022Survey,
  author    = {Wang, R. and Lei, T. and Cui, R. and Zhang, B. and Meng, H. and Nandi, A. K.},
  title     = {Medical image segmentation using deep learning: A survey},
  journal   = {IET Image Processing},
  volume    = {16},
  number    = {5},
  pages     = {1243--1267},
  year      = {2022}
}

@article{moeskops2016automatic,
  title={Automatic segmentation of MR brain images with a convolutional neural network},
  author={Moeskops, Pim and Viergever, Max A and Mendrik, Adri{\"e}nne M and De Vries, Linda S and Benders, Manon JNL and I{\v{s}}gum, Ivana},
  journal={IEEE transactions on medical imaging},
  volume={35},
  number={5},
  pages={1252--1261},
  year={2016},
  publisher={IEEE}
}

@inproceedings{Ronneberger2015UNet,
  author    = {Ronneberger, O. and Fischer, P. and Brox, T.},
  title     = {U-net: Convolutional networks for biomedical image segmentation},
  booktitle = {Medical Image Computing and Computer-Assisted Intervention–MICCAI 2015},
  volume    = {9351},
  pages     = {234--241},
  year      = {2015},
  publisher = {Springer}
}

@inproceedings{Oktay2018AttentionUNet,
  author    = {Oktay, O. and Schlemper, J. and Folgoc, L. L. and Lee, M. and Heinrich, M. and Misawa, K. and Rueckert, D.},
  title     = {Attention U-net: Learning where to look for the pancreas},
  booktitle = {arXiv preprint arXiv:1804.03999},
  year      = {2018}
}

@article{Zhang2018DRUNet,
  author    = {Zhang, Z. and Liu, Q. and Wang, Y.},
  title     = {Road extraction by deep residual u-net},
  journal   = {IEEE Geoscience and Remote Sensing Letters},
  volume    = {15},
  number    = {5},
  pages     = {749--753},
  year      = {2018}
}

@inproceedings{Cicek2016UNet3D,
  author    = {Çiçek, Ö. and Abdulkadir, A. and Lienkamp, S. S. and Brox, T. and Ronneberger, O.},
  title     = {3D U-Net: learning dense volumetric segmentation from sparse annotation},
  booktitle = {MICCAI 2016: Proceedings, Part II},
  pages     = {424--432},
  year      = {2016},
  publisher = {Springer}
}

@article{Azad2024UNetReview,
  author    = {Azad, R. and Aghdam, E. K. and Rauland, A. and Jia, Y. and Avval, A. H. and Bozorgpour, A. and Merhof, D.},
  title     = {Medical image segmentation review: The success of U-net},
  journal   = {IEEE Transactions on Pattern Analysis and Machine Intelligence},
  year      = {2024}
}

@article{Punn2022Survey,
  author    = {Punn, N. S. and Agarwal, S.},
  title     = {Modality specific U-Net variants for biomedical image segmentation: a survey},
  journal   = {Artificial Intelligence Review},
  volume    = {55},
  number    = {7},
  pages     = {5845--5889},
  year      = {2022}
}

@article{Liu2020SurveyUnet,
  author    = {Liu, L. and Cheng, J. and Quan, Q. and Wu, F. X. and Wang, Y. P. and Wang, J.},
  title     = {A survey on U-shaped networks in medical image segmentations},
  journal   = {Neurocomputing},
  volume    = {409},
  pages     = {244--258},
  year      = {2020}
}

@article{Henkelman1985MR,
  author    = {Henkelman, R. M.},
  title     = {Measurement of signal intensities in the presence of noise in MR images},
  journal   = {Medical Physics},
  volume    = {12},
  number    = {2},
  pages     = {232--233},
  year      = {1985}
}

@article{Atanassov1986IFS,
  author    = {Atanassov, K. T.},
  title     = {Intuitionistic fuzzy sets},
  journal   = {Fuzzy Sets and Systems},
  volume    = {20},
  number    = {1},
  pages     = {87--96},
  year      = {1986}
}

@article{Zadeh1965Fuzzy,
  author    = {Zadeh, L. A.},
  title     = {Fuzzy sets},
  journal   = {Information and Control},
  volume    = {8},
  number    = {3},
  pages     = {338--353},
  year      = {1965}
}

@article{kumar2019modified,
  title={A modified intuitionistic fuzzy c-means clustering approach to segment human brain MRI image},
  author={Kumar, Dhirendra and Verma, Hanuman and Mehra, Aparna and Agrawal, RK},
  journal={Multimedia Tools and Applications},
  volume={78},
  number={10},
  pages={12663--12687},
  year={2019},
  publisher={Springer}
}

@article{Verma2025KIFECM,
  author    = {Verma, D. and Verma, H. and Tiwari, P. K.},
  title     = {A hybrid approach for MRI brain image segmentation using KIFECM-IPSO algorithm},
  journal   = {Expert Systems with Applications},
  volume    = {268},
  pages     = {126239},
  year      = {2025}
}

@inproceedings{Price2019FuzzyLayers,
  author    = {Price, S. R. and Price, S. R. and Anderson, D. T.},
  title     = {Introducing fuzzy layers for deep learning},
  booktitle = {2019 IEEE International Conference on Fuzzy Systems (FUZZ-IEEE)},
  pages     = {1--6},
  year      = {2019},
  publisher = {IEEE}
}

@inproceedings{Sharma2019Pooling,
  author    = {Sharma, T. and Singh, V. and Sudhakaran, S. and Verma, N. K.},
  title     = {Fuzzy based pooling in convolutional neural network for image classification},
  booktitle = {2019 IEEE International Conference on Fuzzy Systems (FUZZ-IEEE)},
  pages     = {1--6},
  year      = {2019},
  publisher = {IEEE}
}

@article{Ding2021Infant,
  author    = {Ding, W. and Abdel-Basset, M. and Hawash, H. and Pedrycz, W.},
  title     = {Multimodal infant brain segmentation by fuzzy-informed deep learning},
  journal   = {IEEE Transactions on Fuzzy Systems},
  volume    = {30},
  number    = {4},
  pages     = {1088--1101},
  year      = {2021}
}

@article{Huang2021BreastSeg,
  author    = {Huang, K. and Zhang, Y. and Cheng, H. D. and Xing, P. and Zhang, B.},
  title     = {Semantic segmentation of breast ultrasound image with fuzzy deep learning network and breast anatomy constraints},
  journal   = {Neurocomputing},
  volume    = {450},
  pages     = {319--335},
  year      = {2021}
}

@article{Badawy2021Breast,
  author    = {Badawy, S. M. and Mohamed, A. E. N. A. and Hefnawy, A. A. and Zidan, H. E. and GadAllah, M. T. and El-Banby, G. M.},
  title     = {Automatic semantic segmentation of breast tumors in ultrasound images based on combining fuzzy logic and deep learning—A feasibility study},
  journal   = {PLOS One},
  volume    = {16},
  number    = {5},
  pages     = {e0251899},
  year      = {2021}
}

@article{Huang2022Trustworthy,
  author    = {Huang, K. and Zhang, Y. and Cheng, H. D. and Xing, P.},
  title     = {Trustworthy breast ultrasound image semantic segmentation based on fuzzy uncertainty reduction},
  journal   = {Healthcare},
  volume    = {10},
  number    = {12},
  pages     = {2480},
  year      = {2022}
}

@article{Subhashini2022ThreeWay,
  author    = {Subhashini, L. D. C. S. and Li, Y. and Zhang, J. and Atukorale, A. S.},
  title     = {Integration of fuzzy logic and a convolutional neural network in three-way decision-making},
  journal   = {Expert Systems with Applications},
  volume    = {202},
  pages     = {117103},
  year      = {2022}
}

@article{Chen2022Pancreas,
  author    = {Chen, Y. and Xu, C. and Ding, W. and Sun, S. and Yue, X. and Fujita, H.},
  title     = {Target-aware U-Net with fuzzy skip connections for refined pancreas segmentation},
  journal   = {Applied Soft Computing},
  volume    = {131},
  pages     = {109818},
  year      = {2022}
}

@article{Nan2023FuzzyAttention,
  author    = {Nan, Y. and Del Ser, J. and Tang, Z. and Tang, P. and Xing, X. and Fang, Y. and Yang, G.},
  title     = {Fuzzy attention neural network to tackle discontinuity in airway segmentation},
  journal   = {IEEE Transactions on Neural Networks and Learning Systems},
  year      = {2023}
}

@article{Bustince2000Generators,
  author    = {Bustince, H. and Kacprzyk, J. and Mohedano, V.},
  title     = {Intuitionistic fuzzy generators application to intuitionistic fuzzy complementation},
  journal   = {Fuzzy Sets and Systems},
  volume    = {114},
  number    = {3},
  pages     = {485--504},
  year      = {2000}
}

@article{Yager1980Fuzziness,
  author    = {Yager, R. R.},
  title     = {On the measure of fuzziness and negation. II. Lattices},
  journal   = {Information and Control},
  volume    = {44},
  number    = {3},
  pages     = {236--260},
  year      = {1980}
}

@article{zhou2025f2cau,
  title={F2CAU-Net: A dual fuzzy medical image segmentation cascade method based on fuzzy feature learning},
  author={Zhou, Tianyi and Wang, Haipeng and Geng, Sheng and Ju, Hengrong and Huang, Jiashuang and Fu, Fan and Ding, Weiping},
  journal={Applied Soft Computing},
  pages={113692},
  year={2025},
  publisher={Elsevier}
}

@article{zheng2024systematic,
  title={A systematic survey of fuzzy deep learning for uncertain medical data},
  author={Zheng, Yuanhang and Xu, Zeshui and Wu, Tong and Yi, Zhang},
  journal={Artificial intelligence review},
  volume={57},
  number={9},
  pages={230},
  year={2024},
  publisher={Springer}
}

@article{zhang2025fuzzy,
  title={Fuzzy attention-based deep neural networks for acute lymphoblastic leukemia diagnosis},
  author={Zhang, Tairan and Xue, Gang},
  journal={Applied Soft Computing},
  volume={171},
  pages={112810},
  year={2025},
  publisher={Elsevier}
}

@article{jafrasteh2024enhanced,
  title={Enhanced spatial Fuzzy C-Means algorithm for brain tissue segmentation in T1 images},
  author={Jafrasteh, Bahram and Lubi{\'a}n-Guti{\'e}rrez, Manuel and Lubi{\'a}n-L{\'o}pez, Sim{\'o}n Pedro and Benavente-Fern{\'a}ndez, Isabel},
  journal={Neuroinformatics},
  volume={22},
  number={4},
  pages={407--420},
  year={2024},
  publisher={Springer}
}

@article{noorizadeh2024subject,
  title={Subject-specific atlas for automatic brain tissue segmentation of neonatal magnetic resonance images},
  author={Noorizadeh, Negar and Kazemi, Kamran and Taji, Seyedeh Masoumeh and Danyali, Habibollah and Aarabi, Ardalan},
  journal={Scientific Reports},
  volume={14},
  number={1},
  pages={19114},
  year={2024},
  publisher={Nature Publishing Group UK London}
}

@article{soloh2024brain,
  title={Brain Tumor Segmentation Based on $\alpha$-Expansion Graph Cut},
  author={Soloh, Roaa and Alabboud, Hassan and Shahin, Ahmad and Yassine, Adnan and El Chakik, Abdallah},
  journal={International Journal of Imaging Systems and Technology},
  volume={34},
  number={4},
  pages={e23132},
  year={2024},
  publisher={Wiley Online Library}
}
\end{document}